\documentclass[acmlarge]{acmart}

\AtBeginDocument{%
  }

\acmJournal{POMACS}
\acmVolume{37}
\acmNumber{4}
\acmArticle{111}
\acmMonth{8}

\usepackage{color} 

\begin{document}

\title{Explainable Artificial Intelligence for Medical Applications: A Review}

\author{Qiyang Sun}
\email{q.sun23@imperial.ac.uk}
\orcid{0009-0001-9228-4543}
\affiliation{%
  \institution{Imperial College London}
  \city{London}
  \country{United Kingdom}
  \address{Huxley Building, 180 Queen's Gate, South Kensington, London SW7 2AZ}
}
\author{Alican Akman}
\email{a.akman21@imperial.ac.uk}
\orcid{0000-0002-8010-6897}
\affiliation{%
  \institution{Imperial College London}
  \city{London}
  \country{United Kingdom}
}
\author{Bj{\"o}rn W. Schuller}
\email{bjoern.schuller@imperial.ac.uk}
\orcid{0000-0002-6478-8699}
\affiliation{%
  \institution{Imperial College London}
  \city{London}
  \country{United Kingdom}
}
\additionalaffiliation{%
Technical University of Munich, Munich Data Science Institute, Munich Center for Machine Learning, Munich, Germany
}

\renewcommand{\shortauthors}{Q. Sun et al.}

\begin{abstract}
The continuous development of artificial intelligence (AI) theory has propelled this field to unprecedented heights, owing to the relentless efforts of scholars and researchers. In the medical realm, AI takes a pivotal role, leveraging robust machine learning (ML) algorithms. AI technology in medical imaging aids physicians in X-ray, computed tomography (CT) scans, and magnetic resonance imaging (MRI) diagnoses, conducts pattern recognition and disease prediction based on acoustic data, delivers prognoses on disease types and developmental trends for patients, and employs intelligent health management wearable devices with human-computer interaction technology to name but a few. While these well-established applications have significantly assisted in medical field diagnoses, clinical decision-making, and management, collaboration between the medical and AI sectors faces an urgent challenge: How to substantiate the reliability of decision-making? The underlying issue stems from the conflict between the demand for accountability and result transparency in medical scenarios and the black-box model traits of AI. This article reviews recent research grounded in explainable artificial intelligence (XAI), with an emphasis on medical practices within the visual, audio, and multimodal perspectives. We endeavour to categorise and synthesise these practices, aiming to provide support and guidance for future researchers and healthcare professionals.
\end{abstract}




\received{26 June 2024}
\received[revised]{}
\received[accepted]{}

\maketitle

\section{Introduction}
With the breakthrough of many technical bottlenecks, artificial intelligence (AI) has given rise to several pivotal branches, including deep learning (DL), computer vision (CV), natural language processing (NLP) and large language models (LLMs) \cite{xu2021artificial}. These sub-branches intricately connect, collectively propelling the comprehensive development of AI. As technology continues its relentless evolution, AI has permeated diverse domains, such as education, transportation, and healthcare \cite{mou2019artificial} progressively. 

Recently, AI-based healthcare has been rapidly expanding, leading to the creation of various new techniques for disparate clinical disciplines. The emergence of DL indicated a significant change in the digital technology paradigm, considerably enhancing the precision of healthcare model predictions \cite{miotto2018deep}. For instance, models based on  convolutional neural networks (CNNs) for medical imaging have displayed exceptional accuracy in tasks like tumour identification, organ segmentation, and abnormality detection \cite{xie2021survey,wu2020classification}. On the other hand, recurrent neural networks (RNNs) notably improve voice recognition accuracy, supplanting the traditional GMM-HMM model. This enhancement is supported by the proven performance of RNNs on audio data \cite{yu2017recent}. Consequently, according approaches are being used in acoustic pathology detection systems and telemedicine \cite{pan2012investigation,habib2021toward}. Moreover, the attention-based Transformer architecture is widely used in the research community \cite{vaswani2017attention, valanarasu2021medical}. For example, it is applied to process multimodal medical data, integrating images data along with other medical data such as audio or physiological parameters to obtain a more comprehensive view of health \cite{sun2021multi, abiyev2024multimodal}.

However, the complexity of cases and the massiveness of data render the medical field full of great challenges. 
According to the prediction from International Data Corporation, the global data volume is expected to grow from 33 zettabytes in 2018 to 175 zettabytes by 2025, with healthcare data projected to experience the fastest growth among industries due to advancements in healthcare analytics and increasing frequency and resolution of medical imaging \cite{reinsel2018data}. These vast data sets encompass not only individual information, physiological parameters, and treatment data tailored for patients but also include disease characteristics, finance considerations and cultural nuances \cite{wang2020big,keij2021makes}. Furthermore, the occurrence of unforeseen epidemic diseases, exemplified by the unforeseen COVID-19 pandemic in late 2019, has significantly heightened the strain on the medical system. As of January 28, 2024, global government agencies have validated 774,469,939 reported COVID-19 cases \cite{mathieu2020coronavirus}, with an incalculable death toll attributable to the disease and its complications.
This disaster exposed the problems of shortages of medical personnel, inefficient medical decision-making, and insufficient medical facilities in the medical system \cite{moynihan2021impact}. Therefore, finding effective solutions towards these challenges in such a complex and ever-changing environment is becoming an urgent need in the healthcare field.

Researchers typically engage in exhaustive training of complex and often opaque machine learning (ML) algorithms, iterating many thousands of times to achieve satisfactory outcomes \cite{de2018algorithmic}. Such complex AI models have proven effective in driving innovation and improving model accuracy \cite{holzinger2017we, papenmeier2022s}. Nonetheless, new questions regarding how to elucidate AI model decisions arise: the opaque nature of AI models renders people incapable of comprehending or expounding on the AI system's decision-making process \cite{hoffman2018metrics}. This is primarily due to the fact that several complicated ML models, particularly DL models, involve millions of parameters and hierarchies, contributing to a decision-making process that is intricate and arduous to explain \cite{samek2019explainable}. In the medical domain, however, transparency and explainability in medical decision-making are precisely what doctors and patients need most \cite{holzinger2017we,cummins2020machine}. 

Currently, there are several established explainable AI (XAI) techniques that have been applied and extended within the medical sector \cite{ribeiro2016should,lundberg2017unified,zhou2016learning}. In this article, recent explainability outcomes in medical vision, audio, and multimodal solutions are presented. These achievements entail the evolution of mature XAI models and novel endeavours to enhance explainability.

Several notable reviews explore the intersection of XAI and healthcare. Loh et al.\ \cite{loh2022application} review the applications of XAI in healthcare from 2011 to 2022, highlighting different XAI techniques across various medical settings. However, their review covers a broad timespan and many traditional AI methods they discuss are no longer in use. Furthermore, they do not deeply explore XAI applications related to low-dimensional biosignals. Band et al.\ \cite{band2023application} examine the use of XAI in healthcare, presenting some common XAI methods and how they can be applied to explain specific diseases. They also provide a brief evaluation of implementation approaches. However, the article lacks an in-depth discussion of XAI classification frameworks and does not analyse a wider range of medical XAI studies. Another review by Singh et al.\ \cite{singh2020explainable} focuses on explainable DL models for medical image analysis. Their review concentrates primarily on medical imaging, without considering other data modalities. Similarly, Chaddad et al.\ \cite{chaddad2023survey} review XAI in healthcare, categorising and summarising XAI types and algorithms for medical imaging. However, they cover only a limited number of XAI techniques and do not explore their broader clinical applications.

In contrast, our review makes different contributions,  offering a more comprehensive and novel perspective. We begin by addressing the specific explainability needs within healthcare, highlighting the shared desire for interpretability from both patients and clinicians. We then introduce the definitions, related terms, and taxonomy criteria for XAI. Notably, we classify the explainability needs at the medical level, clarify related terms, and explain their relationships. For taxonomy, we combine the frameworks proposed by \cite{tjoa2020survey}, \cite{holzinger2017we},  \cite{combi2022manifesto}, and \cite{lundberg2020local}, categorising 19 commonly used and promising XAI techniques in healthcare based on four criteria. Furthermore, we analyse over 100 papers published in the past five years that focus on XAI applications across different modalities (vision, audio, and multimodal). We critically evaluate these studies, identify current challenges, and provide outlook for future research directions and the development of XAI applications.

Based on the contributions, the organisational framework of this article is structured as follows. Section \ref{need} elucidates the importance of explanation in the medical field and  the distinct background of XAI within the medical domain. Section \ref{xai} introduces the definition, related terms , taxonomy criteria and detailed technologies of XAI. Section \ref{app}  delves into the advancements of recent medical applications focused on explainability in the realms of visual, audio, and multimodal solutions, respectively. By presenting and thoroughly analysing these applications, Section \ref{discuss} and \ref{out}  discuss the challenges and outlook of XAI within the context of medical applications. Ultimately, Section \ref{con} furnishes a comprehensive summary.

\section{The need for Explanation in Healthcare AI}
\label{need}
Introducing AI models into medical contexts enhances the efficiency and quality of healthcare \cite{Shaban-Nejad2018Health}. However, the lack of explainability in AI models leads to a negative impact on both patients and healthcare professionals.

\subsection{Patients' perspective}

AI medical models enhance the accuracy and precision of medical procedures, yet no model or product can guarantee 100\% accuracy. Negative impacts of errors in AI algorithms are presented in \cite{doi/10.2861/568473}. It is often unclear who bears responsibility for medical accidents that are caused by machine errors. Should it be the doctor, the hospital, the AI product manufacturer, the programmer who wrote the code, or the ones who collected and curated the data used in training? 
This is particularly problematic since opaque AI models make it difficult to assign blame after problems arise, which ultimately fails to protect patients.

Patients who lack medical knowledge face challenges in comprehending the basis for diagnosis results and treatment recommendations. As a result, it is difficult for them to fully participate in medical decision-making. A recent report, 2023 GP Patient Survey held by NHS England and IPSOS with 760,000 respondents \cite{GPPS2023} indicates that only an estimated 60\% of participants felt fully involved in medical decision-making, whilst over 40\% felt they were only involved to some extent or not at all. In 2022, 10\% of patients assumed they did not feel involved in medical decision-making, which is a significant issue. It becomes challenging for doctors and patients to establish complete faith in one another when patients face obstacles while trying to engage in medical decision-making.

In addition, while processing sensitive patient data in medical scenarios, the AI model needs to observe human ethics and legal constraints \cite{savulescu2015moral}. It is crucial to guarantee that patients' private data is not unlawfully misused or leaked upon its submission to the machine.

\subsection{Doctors' perspective}

Models without transparency and explainablility complicate doctors' ability to understand the rationale and inner workings of their decisions, increasing user confusion \cite{Markus2020The}. Doctors are required to thoroughly evaluate the rationale behind cases, diagnosis, and treatment decisions. However, the opacity of complex models can obscure and complicate this evaluative process. Such confusion may lead to doctors doubting the reliability of model recommendations and consequently reducing the integration of AI tools.
    
In the medical domain, AI models frequently use intricate algorithms and mathematical operations, which may be challenging for doctors without technical expertise to understand, thereby limiting their professional development and the advancement of medical practice. Additionally, the millions of parameters in current deep models together with according amounts of data used in training render AI models black boxes at first.

In general, while the integration of AI models into medical practice has certainly improved healthcare efficiency and quality, models that lack explainability present significant challenges for patients and doctors. Hence, it is crucial to guarantee the explainability of healthcare AI, which can foster trust, promote patient engagement, uphold ethical standards, and support physicians in providing high-quality care.

\section{Explainable Artificial Intelligence}
\label{xai}
The improvement of computer explainability dates back to the early days of expert systems. The explainable expert systems(EES) framework, backed by Defense Advanced Research Projects Agency (DARPA) research, recorded essential design features to facilitate generating quality explanations \cite{swartout1991explanations}. Forrest and Mitchell have explored the issue of the explainability of genetic algorithms (GA) by considering the factors that affect the likelihood of success in GA performance \cite{forrest1993makes}.

In the mid-2000s, the term XAI was first introduced, which focuses on the military simulation aspect of generating explainable command models for small unit tactical systems. Unfortunately it did not cause a stir at the time \cite{van2004explainable,lane2005explainable}. The reason is that the majority of research had emphasised on AI algorithm innovation and optimisation while neglecting the explainability of models \cite{minh2022explainable}. In 2016, the \textit{Third Wave of Artificial Intelligence Program} hosted by DARPA incorporated XAI and developed a toolkit library of ML and human-machine interface software modules \cite{gunning2019darpa}. At the same time, a workshop named \textit{The Myth of Model Interpretability} introduced the term XAI, emphasising the growing demand for research on model interpretability in the ML community \cite{lipton2018mythos}.

\begin{figure}
    \centering
  \includegraphics[width=1\linewidth]{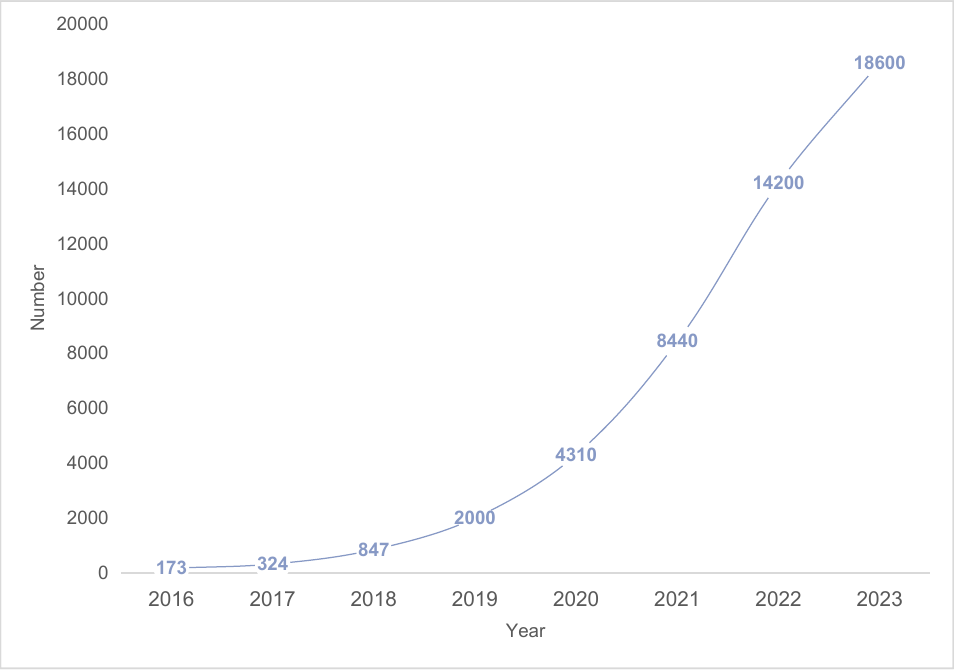}
    \caption{Overview of XAI-related publications retrieved from Google Scholar. Data collected on December 26, 2023, using the search keywords XAI or Explainable Artificial Intelligence.}
    \label{fig:chart0}
    \Description{XAI relate literature trends}
\end{figure}

In recent years, the increasing integration of AI products in various domains of human existence has led to a rise in ethical, moral, and legal demands from society \cite{savulescu2015moral,baird2020considerations,mittelstadt2019principles}. Mere accuracy of results no longer satisfies people, who now expect AI systems to furnish a full and coherent chain of logical justifications \cite{leslie2019understanding}. Such assurances serve to boost their trust in machines. In Fig.~\ref{fig:chart0}, Google Scholar statistics indicate a marked upsurge in research on XAI and related subjects since 2016.

\subsection{Definition}
Academics lack a unified definition of XAI and sometimes conflate interpretability with explainability. Some scholars have attempted to distinguish between these seemingly interchangeable terms, though we are not reiterate these distinctions in this paper \cite{lipton2018mythos,rudin2019stop, broniatowski2021psychological}. Based on the frequency of academic citations, the prevailing definition of XAI is: 

\textit{\textbf{``A set of methods applied to an AI model or model predictions that help explain why the AI made certain decision'' \cite{gunning2019darpa}.}}

There are other definitions of XAI. For instance, Arrieta et al.\ \cite{arrieta2020explainable} define XAI as ``producing details or reasons to make its functioning clear or easy to understand,” focusing on the interpretability of the AI’s internal mechanics. Meanwhile, Minh et al.\ \cite{minh2022explainable} emphasise that XAI is concerned with the ``explainability and transparency for sociotechnical systems, including AI,” highlighting a broader, system-wide perspective. But from our perspective, the definition in \cite{gunning2019darpa} better aligns with the specific needs of medical contexts, where understanding the rationale behind individual AI-driven decisions is often dominant for ensuring trust, safety, and accountability.

To formalise the overarching objective of XAI, consider a general AI model \( f: \mathbb{R}^n \rightarrow \mathbb{R} \) that takes an input \( x \in \mathbb{R}^n \) and produces an output \( y = f(x) \). The goal of XAI is to enhance the explainability of \( f \) by constructing an auxiliary function \( g \), which provides insights into how features in the input \( x \) impact the output \( y \). Specifically, \( g \) is designed to clarify how input features contribute to the output \( y \), either by quantifying the importance of each feature or by revealing other aspects of the model's internal dynamics. This process can generally be expressed as:
\begin{equation}
z = \text{Explain}(f, x),
\end{equation}
where \( \text{Explain}(\cdot) \) is an operator that generates interpretable information about \( f(x) \), with the specific form of \( z \) depending on the particular XAI method employed. While different XAI techniques may represent \( z \) in various ways such as relevance scores and important instances, the common objective remains the same: to provide explanations into how the model \( f \) generates its predictions.

Whilst some literature may use the terms ``explainable AI” (XAI) and ``explainable ML” (XML) interchangeably, XAI encompasses a broader scope, incorporating not only ML but also traditional AI technologies, such as rule-based and expert systems. Presently, ML constitutes the most prominent branch within XAI due to its extensive application and rapid development. For the sake of clarity and consistency, this paper employs the term ``XAI” to denote the comprehensive suite of explainable techniques spanning the entire AI domain.

\begin{figure}
    \centering
  \includegraphics[width=.6\linewidth]{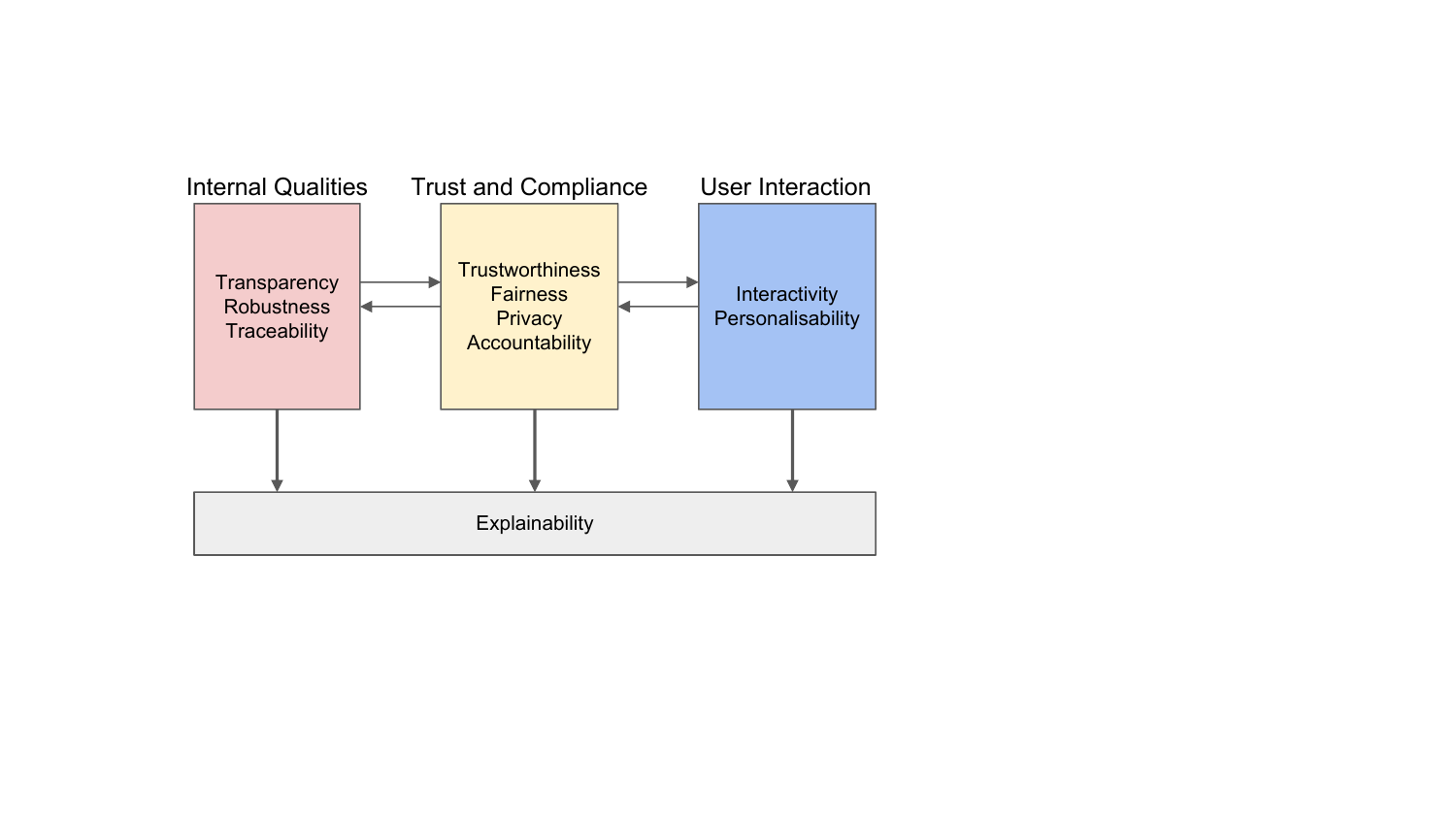}
    \caption{The interrelationship among XAI-related terms}
    \label{fig:chart4}
    \Description{xai flow}
\end{figure}

\subsection{Related Terms}
\label{terms}
The definition elucidates the aim of XAI, which is to facilitate the system in making decisions that align with human logic and to aid humans in creating a comprehensive chain of clarification. In the medical field, explainability can directly impact trust, patient safety, and ethical standards. Consequently, we collect the XAI-relevant terms that are most related to the healthcare context to help make the concept of explainability more accessible.  We divide them into three perspectives: Internal Qualities, Trust and Compliance Factors, and User Interaction Features.

{\textbf{Internal Qualities}}

This category focuses on the internal workings of the model when making decisions.

\textit{Traceability} emphasises that the XAI system can trace a series of steps before the model makes a decision, showing the key features or steps that affect the model's output \cite{rai2020explainable}. 

\textit{Transparency} underscores that the primary aim of XAI is to increase the visibility of AI system processes, enabling users to gain insight into the model's internal workings, understand how decisions are made, and ultimately gain comprehensible results \cite{meske2020transparency}.

\textit{Robustness} is used to describe a system's capacity to maintain consistent performance when confronted with varying inputs. This quality ensures that the model's output remains relatively stable, even in the presence of uncertainty \cite{ali2023explainable}.
 
\textbf{Trust and Compliance}

This category concentrates on the system's performance in terms of medical ethics, data privacy, and trust.

\textit{Trustworthiness} demands that XAI improves user trust in the AI system by guaranteeing a dependable decision-making process that is not adversely influenced \cite{minh2022explainable}. 

\textit{Fairness} necessitates the XAI system to promote just decisions but also just explanations towards diverse groups, without reinforcing or amplifying biases in the data \cite{angerschmid2022fairness}.

\textit{Privacy} prioritises the XAI system's responsibility to safeguard user data and ensure that personal information remains confidential whilst providing explainability methods \cite{baird2020considerations}.

\textit{Accountability} denotes the ability of the XAI to understand why an error occurred after the system has made an error or encountered an unforeseen circumstance, which in turn can assist in the question of attributing responsibility for the accident and explaining it \cite{arrieta2020explainable}.

\textbf{User Interaction}

This category addresses how the system interacts with healthcare professionals and patients to meet their explanatory needs.

\textit{Interactivity} centres on the XAI model's capacity to engage with the user, with the intent to offer clarification on any confusion arising from the interaction process, ultimately enhancing the user's comprehension of the system \cite{wang2021explainability}.

\textit{Personalisability} stands for the ability of an XAI system to adapt its explanations to the user, such as towards
their demographic data, medical history, modality and modality combination preference and furthermore \cite{nimmo2024user}.

Figure \ref{fig:chart4} illustrates the interrelationships among those related terms. The internal qualities of the model ensure the reliability and comprehensibility of its internal processes, allowing the system to perform in a trustworthy and consistent manner. Simultaneously, user interaction factors are essential for meaningful engagement with healthcare professionals and patients. These features enable users to seek clarification, ask questions, and receive tailored explanations that match their specific needs and levels of understanding. Together, these two elements directly influence trust and compliance, establishing a vital connection between the AI system and healthcare users. All of these factors contribute directly to explainability. Ultimately, these related terms underscore that explainability encompasses not only an understanding of how a model functions but also the development of a powerful, trusted, and interactive system that can effectively address the demands of healthcare environments.

\subsection{Taxonomy Criteria}
\label{criteria}
Scholars have proposed various perspectives to categorise XAI methods. In \cite{tjoa2020survey}, XAI techniques are separated into two primary categories according to human perception: perceptive interpretability and interpretability via mathematical structures. In \cite{holzinger2017we}, two main categories of XAI techniques are distinguished based on the order in which explanations are provided: ante-hoc models and post-hoc models. In \cite{combi2022manifesto}, XAI techniques are divided into two categories based on the characteristics of the target model:  model-agnostic approaches and model-specific approaches. \cite{lundberg2020local} classified XAI techniques based on the local and global nature of the explanations. In addition, a more intricate taxonomy or combination of approaches is necessary to fully describe the diversity of XAI techniques \cite{schwalbe2023comprehensive}.

\textbf{From the human perception perspective}

\textit{Perceptive Interpretability}: Techniques generated by such interpretive methods are usually considered to be immediately interpretable. This means that the explanations generated by these methods can be directly understood by humans without additional cognitive processing. For example, some visualisation techniques (e.\,g., feature importance maps, decision trees) provide perceptual interpretability. These methods typically provide easily understandable explanations by highlighting the significance of input features, often visualised, or by observing the stimulation of neurons or groups of neurons, known as signal interpretability.

\textit{Interpretability by Mathematical Structures}: In mathematical logic, interpretability is a relation that expresses the possibility of interpreting or translating one theory into another. These interpretive methods provide explanations through mathematical structures. The output generated needs to go through an additional layer of cognitive processing interfaces to reach an interpretable interface. Explanations generated by these methods often require mathematical knowledge to be understood. For instance, some ML algorithms use explanatory mechanisms based on mathematical structures, such as logistic regression, generative discriminators, and reinforcement learning.

\textbf{From the order perspective}
 
\textit{Ante-hoc Model}: The term ``ante-hoc'' originates from the Latin word ``ante'', meaning ``before''. Ante-hoc models are designed and trained with interpretability in mind, often sacrificing some predictive performance for better interpretability. They provide insights into model behaviour at an early stage of model training. However, this approach may limit the complexity and predictive power of the model.

\textit{Post-hoc Model}: ``Post-hoc'' also stems from Latin meaning ``after this”. The goal of post-hoc models is to provide explainability after model training is complete. They are often applied to explain complex models that are considered ``black boxes”, such as DL models. These methods approximate the internal workings of a model so that it can be understood by humans. They can be used to explain any type of model, including those that are very complex. However, this approach has a major disadvantage: the explanations it generates can be difficult to understand because of the model's complexity.

\textbf{From the model perspective}

\textit{Model-agnostic Approach}: A model-agnostic approach is an approach to generate explanations of a model's decision process without relying on the internal structure of a particular model. The approach considers AI models as black boxes and provides explanations without any prior knowledge. Therefore, it can be applied to any AI model.

\textit{Model-specific Approach}: A model-specific approach is a method that relies on the detailed structure of a particular ML or DL model. This approach is commonly referred to as the intrinsic approach because it allows for access to the model's internal structure. A detailed explanation of the decision-making process requires a thorough understanding of the model. In contrast to the black-box testing approach of model agnostic methods, model-specific methods operate more similarly to the white-box testing of software applications.

\textbf{From the coverage perspective}

\textit{Local Explanation}: A local explanation is concerned with explaining the decisions made by the model in individual instances. These methods are often used to explain complex models that are considered `black boxes'. They explain the inner workings of a model by approximating its logic so that it can be understood by a human examiner. Local explanations can only explain a single prediction and are not valid for the general behaviour of the model.

\textit{Global Explanation}: A global explanation attempts to answer the question of how the parameters of the model affect its decisions. It tries to explain the overall behaviour of the model, not just the individual predictions. Therefore, achieving global explainability may be difficult when the model has many parameters.

These taxonomy criteria are not irrelevant to each other. For example, in some literature, model-agnostic approaches and model-specific approaches appear as subcategories of post-hoc models \cite{gohel2021explainable}. Further, many more could be added, such as the modalities perspective, i.\,e., by which modality or combination of such the explanation is given, including visualisation, sonification, or verbalisation amongst other. 
Additionally, numerous recent review articles have advocated for the integration of diverse taxonomy criteria to achieve a comprehensive assessment of XAI technology \cite{xie2021survey,gohel2021explainable,agarwal2020interpretable,van2022explainable,vilone2020explainable}. This multifaceted approach enables the evaluation of XAI methods from various perspectives, thereby facilitating a deeper understanding of their practical effectiveness and allowing for targeted recommendations tailored to the specific needs of distinct domains and user groups. This article focuses specifically on the medical field, where XAI technology is assessed through this integrative, multi-criteria framework to meet the sector’s unique demands.

\subsection{General XAI Techniques}
\label{tech}
19 XAI techniques were identified through a comprehensive literature review of medical XAI applications. These selected techniques include commonly used XAI methods in medical applications (e.\,g., LIME and Grad-CAM) and promising approaches (e.\,g., RISE and CRP), each rooted in a robust theoretical foundation. They are classified into five categories based on their implementation principles: 1) Perturbation-based approaches; 2) Backpropagation-based approaches; 3) Gradient-based approaches; 4) Instance-based approaches and 5) Other approaches.

\textbf{Perturbation-based approaches}

Perturbation-based approaches focus on assessing the sensitivity of a model's predictions by applying small perturbations to the input features and observing the resulting changes in the output. Let \( f(x) \) denote the model’s prediction for a given input \( x \in \mathbb{R}^n \). For a perturbed input \( x' = x + \delta \), where \( \delta \in \mathbb{R}^n \) is a small perturbation vector, the change in the model output due to this perturbation can be represented as:
\begin{equation}
\Delta y = f(x + \delta) - f(x).
\end{equation}
In this context, \( \Delta y \) measures the influence of the perturbation \( \delta \) on the model output, providing insight into the importance of the perturbed features within \( x \). By systematically varying \( \delta \), different subsets of \( x \) can be perturbed to evaluate the contributions of individual features or groups of features. This process enables an approximation of how sensitive the model is to changes in its input, thereby offering an interpretable view of the model’s decision-making process.

\textit{Local Interpretable Model-agnostic Explanations (LIME)} \cite{ribeiro2016should}  is a widely used perturbation approach in XAI that trains an explainable model, such as linear regression or a decision tree, by generating new sample points around the selected sample points and using these new sample points and the predicted values from the black-box model. A similarity calculation and the features to be selected for explanation are then defined. Sample weights are assigned based on the distance to the sample points after perturbations are sampled around them. This approach allows for obtaining a good local approximation of the black-box model, which means that an explainable model can be used to explain the complex model locally \cite{lee2019developing}. 

\textit{Occlusion Sensitivity Map (OSM)} is proposed in \cite{zeiler2014visualizing} as a method for calculating the importance of an input. This is achieved by placing a grey square over a portion of the input image and measuring the effect of this occlusion on the model prediction. OSM provides an intuitive understanding and explanation of the predictive behaviour of DL models. By sliding the grey square, changes in the model's predictive performance are observed.

\textit{Adversarial Training (AT)} \cite{malik2022xai} involves adding small perturbations to the model's input samples to significantly alter the model's predictions. The model is trained to adapt to these perturbations. Analysing the differences between the adversarial samples and the original samples provides researchers with a better understanding of the model's decision boundaries and allows them to understand and improve the model's robustness \cite{al2022xai}.

The \textit{Randomised Input Sampling For Explanation (RISE)} \cite{petsiuk2018rise} approach is to use Monte Carlo sampling to generate a random binary mask, and then perform a weighted averaging of the random masks, where the weights are the output of the masked model. Through multiple sampling and weighted averaging, a saliency map is eventually generated. The final result characterises the sensitivity of the model to different regions of the input image.

\textbf{Backpropagation-based approaches }

Neural networks use backpropagation algorithms to compute the partial derivatives of the loss function concerning the weights of each neuron. This constitutes the gradient of the objective function concerning the vector of weights, which is then minimised to reduce the loss function \cite{rojas1996backpropagation}. However, not all backpropagation-based algorithms in the field of XAI are focused on calculating the gradient. The paper divides backpropagation-based algorithms and gradient-based methods into two categories, treating them as separate and independent subjects of discussion. Backpropagation-based approaches in XAI utilise the model's backpropagation network to trace the model’s output back to the input layer, thereby attributing contributions to each input feature. Let \( f(x) \) represent the model's prediction for a given input \( x \in \mathbb{R}^n \), and let \( y = f(x) \) denote the model’s output. Using backpropagation, these approaches compute a relevance score \( R_i \) for each input feature \( x_i \), which quantifies the contribution of \( x_i \) to the output \( y \). The relevance \( R_i \) is propagated back through the network using layer-specific propagation rules. The relevance propagation from one layer \( l \) to the preceding layer \( l-1 \) can generally be expressed as:
\begin{equation}
R_i^{(l)} = \sum_j \frac{z_{ij}}{\sum_k z_{kj}} R_j^{(l+1)},
\end{equation}
where \( R_i^{(l)} \) represents the relevance score of the \( i \)-th neuron in layer \( l \); \( z_{ij} \) denotes the connection strength or activation between neuron \( i \) in layer \( l \), and neuron \( j \) in layer \( l+1 \) and \( R_j^{(l+1)} \) are the relevance propagated from the subsequent layer \( l+1 \).

By propagating relevance scores through all layers, backpropagation-based approaches attribute portions of the model’s output \( y \) back to each input feature \( x_i \), thereby revealing the features that highest contribute to the prediction. This process provides an explainable breakdown of how input features influence the model’s decision.

The \textit{Layer-wise Relevance Propagation (LRP)} approach is based on the idea of backpropagation \cite{bach2015pixel}. The goal is to assign relevance scores to each input feature or neuron in the network to indicate its contribution to the output prediction. LRP recursively propagates the relevance scores from the output layer of the network through a Deep Taylor Decomposition (DTD) propagation to the input layer. At each layer, the relevance scores are redistributed to the input neurons based on their contribution to the output activation of that layer. This redistribution is performed using a set of propagation rules that ensure that the sum of the relevance scores at each layer is conserved.

\textit{Concept Relevance Propagation (CRP)} \cite{achtibat2023attribution} is a novel XAI approach that combines local and global perspectives to answer ``where'' and ``what'' questions about individual predictions. CRP does not rely on gradients directly. The model uses a unique backpropagation technique that considers the correlation between features. This technique involves backpropagating the correlation from the output layer to the input layer to determine the contribution of each input pixel to a specific prediction \cite{achtibat2022towards}.

\textit{Class Activation Mapping (CAM)} \cite{yang2019towards} is an approach used to visualise the regions of interest of a CNN for image classification. It provides an intuitive explanation of the model's decisions in the input image. The key idea of CAM is to generate a weighting matrix by looking at the activation values of the last convolutional layer in a convolutional neural network, in conjunction with a Global Average Pooling (GAP) operation. This is followed by an Average Pooling operation to generate a weight matrix. The weight matrix indicates the significance of each pixel position for a specific category. Multiplying this weight matrix with the output of the convolutional layer produces the final CAM, which is the heat map of the region of interest. After years of development and iteration, CAM has spawned many excellent branches and has been divided into two classes. One class does not require the involvement of gradients, which is in the initial idea of CAM and includes methods such as Score CAM \cite{wang2020score}, ss-CAM \cite{wang2020ss} and Ablation CAM \cite{ramaswamy2020ablation}. The other category is gradient-based CAM, which includes Grad-CAM \cite{selvaraju2017grad}, Grad-CAM++ \cite{chattopadhay2018grad}, and Smooth Grad-CAM++ \cite{omeiza2019smooth}, which will be introduced in the gradient-based approach section.

\textit{Attention Mechanism (AM)} \cite{vaswani2017attention} is commonly used to clarify model decisions. It calculates a ``soft” weight for each element in the input data, which can be interpreted as the amount of ``attention” the model pays to each input element when making predictions. These weights are optimised by backpropagation. This mechanism enables the model to provide explainability by focusing on the input elements that have the most significant impact on the prediction when dealing with complex input data. However, some scholars have disputed the explainability provided by the attention mechanism \cite{serrano2019attention}. In this paper, we argue that AM can produce convincing results of  explainability. But this does not mean that all applications based on the transformer architecture are explainable. 

\textbf{Gradient-based approaches}

Gradient-based approaches refer to a class of methods that use gradients to explain decisions or predictions made by ML models. These methods involve calculating gradients with respect to input features to understand how changes in input variables affect the model's output. Let \( f(x) \) represent the model’s output for an input \( x \in \mathbb{R}^n \), where the prediction is denoted as \( y = f(x) \). The gradient of \( f(x) \) with respect to the input \( x \) is given by:
\begin{equation}
\nabla_x f(x) = \left( \frac{\partial f(x)}{\partial x_1}, \frac{\partial f(x)}{\partial x_2}, \dots, \frac{\partial f(x)}{\partial x_n} \right),
\end{equation}
where \( \frac{\partial f(x)}{\partial x_i} \) represents the partial derivative of \( f(x) \) with respect to the \( i \)-th input feature \( x_i \). This gradient vector \( \nabla_x f(x) \) quantifies the sensitivity of the model’s output to each input feature. In this way, gradient-based approaches focus on the direct sensitivity of the model's output to changes in the input features.

\textit{Integrated Gradients (IG)} \cite{sundararajan2016gradients} evaluates the contribution of each input feature to the model's prediction by calculating the integral of the gradient path over the input space. IG allows users to comprehend the relative impact of various features on predictions by illustrating how each feature evolves across the entire gradient path.

As previously stated, Grad-CAM utilises gradient weighting of the last convolutional layer's output to generate more detailed heatmaps, in contrast to CAM which solely relies on global average pooling. This method, which uses gradient weighting, highlights the significance of each pixel's position in the model's prediction. It provides users with a more precise and understandable explanation of the focus strategy of convolutional neural networks in image classification tasks. 

\textit{Deep Learning Important Features (Deep-Lift)} \cite{vsimic2021xai} analyses the contribution of input features by comparing the difference in neuron activation states relative to the baseline state. Deep-Lift utilises the chain rule and gradient information to determine the impact of each input feature on the model output, which provides a comprehensive understanding of the neural network's internal activations and weights, enabling users to explain the model's decision-making process in specific input scenarios.

\textit{Saliency Map (SM)} \cite{simonyan2013deep} generates a map that highlights the relative importance or prominence of each element in the input data (such as pixels of images, and time-frequency domain segments of audio) to the model's decision-making. The SM technique generates a map that highlights the relative importance or prominence of each element in the input data (such as pixels of images, time-frequency domain segments of audio, etc.) to the model's decision-making. This is achieved by calculating the gradient of each element concerning the model output. 

\textit{Guided Backpropagation (GBP)} \cite{springenberg2014striving} technique is used to visualise the impact of the input on the model output by backpropagating gradients. GBP aims to produce a modified gradient map that highlights the input elements that are more important for the model output. In comparison to traditional backpropagation, GBP includes a regularisation step to improve explainability by preserving positive gradients and suppressing negative gradients \cite{lin2021you}.

\textit{Testing with Concept Activation Vectors (TCAV)} \cite{kim2018interpretability} employs directional derivatives to measure the significance of user-defined concepts to the classification results. Unlike other methods that display the importance weight of each input feature, TCAV displays high-level concepts such as colour, gender, and race to predict class importance. This approach provides a more comprehensive understanding of the factors that contribute to the prediction.

It is noteworthy that several studies have emphasised the importance of baseline selection in gradient-based XAI methods. Different baseline choices, such as a zero vector or an average input, can substantially influence the outcomes of feature importance attribution. An appropriate baseline facilitates a more accurate representation of the relative contributions of input features to the model's predictions, while an unsuitable baseline may yield inconsistent explanations \cite{ancona2017towards, wang2024gradient}.

\textbf{Instance-based XAI approaches}

Instance-based XAI approaches primarily focus on the interpretation of a single sample or group of samples. These methods treat the model as a black box, without considering its internal complexity. For a single sample, there may be a linear or monotonic relationship between the predicted value and some features. By analysing the relationships between inputs and outputs, rather than examining the internal workings of the model, instance-based approaches are applicable to any model type. Given an input \( x \in \mathbb{R}^n \) with its corresponding prediction \( y = f(x) \), instance-based approaches aim to identify related instances from a dataset \( S \) that can provide insights into the model's decision for \( x \). This can be generally formalised as:
\begin{equation}
E(x) = \{ x' \in S \mid \text{rel}(x, x') \geq \tau \},
\end{equation}
where \( S \) represents the set of available instances (e.\,g., training data), \( \text{rel}(x, x') \) is a relevance or similarity function that quantifies the relationship between the input instance \( x \) and another instance \( x' \), and \( \tau \) is a threshold that determines the level of similarity or relevance needed for \( x' \) to be considered useful for explaining \( x \).

Instance-based approaches thus provide interpretability by identifying instances that are either similar, contrasting, or exhibit characteristics that help elucidate the model’s decision-making process, offering a context-rich explanation that is model-agnostic and applicable across different ML architectures.

\textit{Multiple Instance Learning (MIL)} \cite{electronics12204323} is a weakly supervised learning method that can improve model explainability. MIL organises data into bags, each containing multiple instances. Although each bag has a label, the instances within the bag do not have specific labels. This structure enables to identify the instances that contribute the most to the bag's label, thereby enhancing model explainability.

\textit{Counterfactual Explanation (CE)} \cite{keane2020good} is a technique used to explain model decisions. It aids in understanding the decision-making process of the model by describing how the model's predictions would change if the input data were altered. The fundamental question of counterfactual explanation is: ``If the input characteristics were to undergo a certain change, how would the output result change?''  Although some scholars consider the relationship between counterfactual explanation and XAI to be controversial, recent research suggests that counterfactual explanation can provide users with a true understanding of the method's support. 

\textit{Anchor} \cite{hagras2018toward} uses an if-then rule to explain the model's predicted behaviour in the target area. Although the original model may be complex and difficult to comprehend, if-then rules are highly explainable and easy for people to understand and follow.

\textbf{Other approaches}

We next describe the ``other'' category of methods that are difficult to classify into the previously mentioned categories and have their characteristics.  These methods are also widely used in the field of XAI.

The \textit{Shapley Additive exPlanations (SHAP)} method \cite{lundberg2017unified} integrates the concept of shapley values from game theory into an additive feature attribution framework. In SHAP, each input feature is treated as a player in a cooperative game, where the predicted output represents the total payoff. To determine each feature's contribution, SHAP calculates Shapley values based on a weighted average of all possible feature permutations, with each feature assigned a weight according to its presence in each permutation. This process yields a shapley value for each feature, indicating its specific contribution to the final prediction. Some papers place SHAP within gradient-based approaches as classification \cite{jin2023guidelines}, but we think while some variants of SHAP may be used in conjunction with other techniques \cite{jia2020explaining,khalane2023evaluating}, the SHAP algorithm itself does not directly involve gradients or perturbations.

The \textit{Partial Dependence Plot (PDP)} \cite{klosok2020towards} is used to visualise the relationship between features and prediction results in a ML model. It shows how the model changes as a single feature changes while holding other features constant. By drawing partial dependency graphs of features, users can intuitively observe the model's response to a single feature and better understand the model's behaviour.

\begin{figure}
    \centering
    \includegraphics[width=\textwidth]{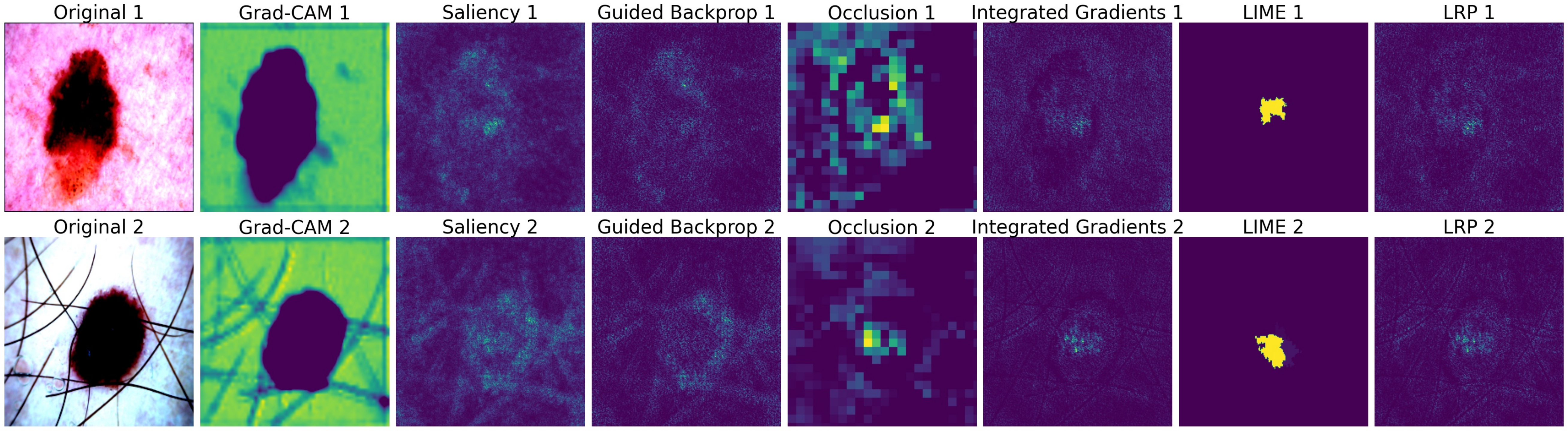}
    \caption{Demonstration of some XAI techniques. Original images are from ISIC2016 Challenge skin lesion datasets \cite{gutman2016skin}.}
    \label{fig:chart3}
    \Description{Demonstration of some XAI techniques}
\end{figure}

\begin{table*}
\begin{center}%
\resizebox{\textwidth}{!}{
\begin{tabular}{ccccccccc}%
\toprule
Techniques & Perception & Mathematics & Ante-hoc & Post-hoc & Model-agnostic & Model-specific & Local & Global \\
\midrule
\textbf{Perturbation-based} & &  &  &  && &&  \\
LIME \cite{ribeiro2016should} &\checkmark  &  &  & \checkmark &\checkmark & &\checkmark &  \\
OSM \cite{zeiler2014visualizing} & \checkmark &  &  & \checkmark &\checkmark & &\checkmark &  \\
AT \cite{malik2022xai} & \checkmark &  & \checkmark &  & &\checkmark & & \checkmark \\
RISE \cite{petsiuk2018rise} &\checkmark  &  &  & \checkmark &\checkmark&  & \checkmark&  \\
\hline
\textbf{Backpropagation-based} & &  &  &  && &&  \\
LRP \cite{bach2015pixel} & \checkmark  &  &  & \checkmark & & \checkmark&\checkmark &  \\
CRP \cite{achtibat2023attribution} & \checkmark   &&  & \checkmark & &\checkmark &\checkmark & \checkmark  \\
CAM \cite{yang2019towards} & \checkmark &  &  & \checkmark & & \checkmark&\checkmark &  \\
AM \cite{vaswani2017attention} & \checkmark &  &   & \checkmark& & \checkmark & \checkmark &  \\
\hline
\textbf{Gradient-based} & &  &  &  && &&  \\
IG \cite{sundararajan2016gradients} & \checkmark &  &  & \checkmark & & \checkmark&\checkmark &  \\
Grad-CAM \cite{selvaraju2017grad} &\checkmark  &  &  & \checkmark & &\checkmark &\checkmark &  \\
Deep-Lift \cite{vsimic2021xai}  & \checkmark &  &  & \checkmark &&\checkmark  &\checkmark &  \\
SM \cite{simonyan2013deep}  &\checkmark  &  &  & \checkmark & & \checkmark  &\checkmark & \checkmark  \\
GBP \cite{springenberg2014striving} & \checkmark &  &  & \checkmark & & \checkmark &\checkmark & \\
TCAV \cite{kim2018interpretability} &  &\checkmark  &  & \checkmark & & \checkmark & & \checkmark \\
\hline
\textbf{Instance-based} & &  &  &  && &&  \\
MIL \cite{electronics12204323} &  & \checkmark & \checkmark &  & \checkmark& & & \checkmark  \\
CE \cite{keane2020good} & \checkmark &  &  & \checkmark &\checkmark & &\checkmark &  \\
Anchor \cite{hagras2018toward} & &  \checkmark &   &  \checkmark&\checkmark & &\checkmark &  \\
\hline
\textbf{Other} & &  &  &  && &&  \\
SHAP \cite{lundberg2017unified} & \checkmark &  &  & \checkmark &\checkmark & &\checkmark &  \\
PDP  \cite{klosok2020towards} & \checkmark &  &  & \checkmark &\checkmark & & &  \checkmark\\
\bottomrule
\end{tabular}
}
\end{center}
\caption{Taxonomy of XAI techniques. XAI techniques are abbreviated as outlined in Section \ref{tech}.}
\label{tab:table1}
\end{table*}

Figure \ref{fig:chart3} lists raw skin lesion images and heat maps obtained using some XAI techniques in a melanoma image classification task. Each technique highlights the areas of the image that have the most significant impact on model decisions. Table \ref{tab:table1} presents the classification results for each technology based on the evaluation criteria outlined in Section \ref{criteria}. 

It is important to note that some technologies may have different initial intentions and varying interpretations by different researchers, leading to discrepancies between papers. For instance, some articles claim that saliency maps are only suitable for local explanations \cite{van2022explainable}, some articles insist that saliency maps are suitable for global explanations \cite{angelov2021explainable}, and many articles emphasise that saliency maps are suitable for both local and global explanations \cite{Wang2020Video}. This controversy is reasonable as there are currently no clear specifications for classifying these methods. This table presents novel and unique research on XAI, covering four different evaluation criteria that have not been included simultaneously in other literature. The contents of table are derived from previous research summaries and our subjective understanding, and therefore some parts may deviate from the classification results of other researchers.

\section{XAI Applications in Medicine Review}
\label{app}
In this section, we assess the visual, audio, and multimodal domains by reviewing selected publications and presenting a comparative analysis in tabular form to highlight current trends. In each subsection, only a few representative studies are summarised by text. We only select studies that offer novel insights into XAI applications with different medical tasks, rather than those that solely utilise XAI techniques without deeper analysis. 

The selection of publications follows a systematic approach, primarily using Google Scholar as the main database.  We combine modality-specific terms (e.\,g., ``visual,” ``audio,” and ``multimodal”) AND core keywords ``XAI” OR ``explainable artificial intelligence" AND ``healthcare” OR ``medicine.” The XAI related terms mentioned in Section \ref{terms} are also considered. This ensures that the selected studies are relevant to the specific domain and healthcare context. To capture the most recent developments, we limit the search to papers published between 2018 and 2024. Publications are chosen based on their relevance to XAI applications in healthcare, with particular attention to studies proposing novel explainability methods, clinical applications, or comprehensive evaluations of XAI models.

\subsection{Visual XAI Applications in Medicine}
In the medical field, XAI is crucial for multidimensional vision scenarios. In one dimension (1D), it can interpret physiological signals related to time series, such as electrocardiogram (ECG) or continuously monitored vital signs. In two dimensions (2D), XAI helps doctors to more accurately interpret 2D images such as X-rays, CT scans, and MRIs. In three dimensions (3D), applications of XAI cover the field of pathology, providing doctors with detailed interpretations of volumetric data (e.\,g., tumour size and shape), helping to achieve accurate diagnosis of early-stage disease and precise treatment plans.

\begin{table*}
\begin{center}
\resizebox{\textwidth}{!}{
\begin{tabular}{cccccc}
\toprule
Name & XAI Techniques & Anatomical Location & Modality & Cons \\
\midrule

Retinoblastoma Diagnosis \cite{aldughayfiq2023explainable} & LIME, SHAP & Eyes & Retinal Fundus Images & Practicality \\ 


Glaucoma Diagnosis \cite{deperlioglu2022explainable} & CAM, Grad-CAM*, AM & Eyes & Retinal Fundus Images & Performance-Explanation Trade-off \\

Fundus Blood Vessel Segmentation \cite{liu2023res2unet} & AM & Eyes & Retinal Fundus Images & Stability\\

Glaucoma Detection \cite{chayan2022explainable} & LIME & Eyes & Retinal OCT & Hardware \\
 
Fungal Keratitis Diagnosis \cite{xu2021clinical} & Grad-CAM* & Eyes & Confocal Microscopy Images & Selection Bias, Generalisability\\ 

COVID-19 Diagnosis \cite{li2022explainable} & MIL, Grad-CAM++ & Chest & CT & Data Compatibility, Explanation Effect\\

COVID-19 Diagnosis \cite{goel2022effect} & Grad-CAM, OSM, LIME & Chest & CT & User Understanding\\

Pneumonia and Tuberculosis Classification \cite{bhandari2022explanatory} & Grad-CAM,LIME,SHAP & Chest & X-ray & Data Compatibility\\

Pneumonia Infection Classification \cite{sheu2023interpretable} & SHAP & Chest & X-ray & Security\\

Pneumonia Classification \cite{madan2023explainable} & Deep-Lift & Chest & CT & Generalisability\\

Lung Disease Detection \cite{prasad2023explainable} & LIME & Chest & X-ray & Interference\\

Pneumonia Detection \cite{mertes2022ganterfactual} & CE & Chest & X-ray & Universality \\

Pneumonia Detection \cite{mridha2023accuracy} & Grad-CAM* & Chest & X-ray & Diversity \\

Pneumonia Identification \cite{ukwuoma2023hybrid} & SM, Grad-CAM & Chest & X-ray & Accuracy\\

COVID-19 Diagnosis \cite{teixeira2021impact} & LIME, Grad-CAM & Chest & X-ray & Database Bias\\

Pneumonia Identification \cite{yang2022deep} & Grad-CAM & Chest & X-ray & Data Scarcity \\

Pneumonia and Tuberculosis Classification \cite{dagnaw2023towards} & Score-CAM & Chest & X-ray & Practicality \\

Pneumonia Classification \cite{dipto2023pnexai} & LIME & Chest & X-ray & Data Limitation \\

Lung Cancer Detection \cite{wani2024deepxplainer} & SHAP & Chest & CT & Accessibility \\ 

Lung Cancer Screening \cite{kobylinska2022explainable} & SHAP & Chest & CT & Stability\\

Lung Abnormalities Detection \cite{islam2023enhancing} & LIME,SHAP,Grad-CAM & Chest & CT & Time Sensitivity\\

Alzheimer's Disease Analysis \cite{jin2020generalizable} & AM & Brain & MRI & Practicality \\ 

Alzheimer's Disease Analysis \cite{sudar2022alzheimer} & LRP & Brain & MRI & Usability\\ 

Brain-Hemorrhage Classification \cite{highton12evaluation} & RISE & Brain & 3D CT & Hardware \\

Alzheimer's Disease Prediction \cite{shad2021exploring} & LIME & Brain & MRI & Accessibility\\ 

Brain Tumour Prediction \cite{gaur2022explanation} & LIME,SHAP & Brain & MRI & Hardware\\ 

Barrett’s Esophagus Classification \cite{de2021convolutional} & SM & Esophagus & OCT & Generalisability \\

Melanoma Recognition \cite{yan2019skin} & AM & Skin & Dermoscopic Images & Binary-Only\\


Skin Lesion Classification \cite{lucieri2020interpretability} & TCAV & Skin & Dermoscopic Images & Data Scarcity \\

Skin Lesion Diagnosis \cite{metta2023improving} & SM & Skin & Dermoscopic images & Time-Consuming \\

Skin Cancer Classification \cite{saarela2022robustness} & IG, LIME & Skin & Dermoscopic Images & Legality \\

Skin Diseases Detection \cite{athina2022multi} & LIME & Skin & Dermoscopic Images & Versatility  \\

Skin Diseases Diagnosis \cite{mayanja2023explainable} & Grad-CAM* & Skin & Clinical Images &  Generalisability\\

Monkeypox Detection \cite{nayak2023detection} & LIME & Skin & Clinical Images & Practicality\\

Skin Cancer Classification \cite{mridha2023interpretable} & Grad-CAM* & Skin & Dermoscopic Images & Generalisability \\

Skin Disease Detection \cite{ballari2022explainable} & Grad-CAM & Skin & Dermoscopic Images & Sensitivity \\

Skin Lesion Classification \cite{corbin2023assessing} & Grad-CAM & Skin & Dermoscopic Images & Flexibility \\

Erythemato-Squamous Diseases Prediction \cite{rathore2022erythemato} & SHAP & Skin & Dermoscopic Images & Practicality \\

Skin Cancer Classification \cite{khater2023skin} & PDP, SHAP & Skin & Dermoscopic Images & Model Limitation \\

Pigmented Skin Lesions Diagnosis \cite{hurtado2022use} & LIME, SHAP & Skin & Dermoscopic Images & Hardware \\

Breast Cancer Diagnosis \cite{chen2021breast} & AM & Breast & Mammography & Complexity\\

Breast Carcinoma Detection \cite{la2023xai} & LRP & Breast & X-ray & Trustworthiness\\

Breast Lesion Classification \cite{hussain2022shape} & Grad-CAM, LIME & Breast & Mammography & Coarseness  \\

Breast Cancer Malignancy Predictions \cite{hartley2023local} & SHAP & Breast & Mammography & Hardware \\

Liver Segmentation \cite{bardozzo2022cross} & Grad-CAM* & Liver & Laparoscopic Images & Sparsity\\

White Blood Cells Monitoring \cite{dipto2023xai} & Grad-CAM & Blood & Microscopic Images & Sparsity \\


White Blood Cells Classification \cite{bhatia2023integrating} & LIME & Blood & Augmented Images & Generalisability \\

Malaria Cell Classification \cite{raihan2022malaria} & SHAP & Blood & Microscopic Images & Comparability  \\

Whole-blood DNA Methylation Classification \cite{kalyakulina2022disease} & SHAP & Blood & Cancer Genome Atlas & Hardware \\

Acute Lymphoblastic Leukemia Classification \cite{diaz2023explainable} & Grad-CAM* & Blood & Microscopic Images & Time-consuming  \\

Kidney Tumor Segmentation \cite{rao2023optimizing} & Grad-CAM & Kidney & CT & Diversity \\


Renal Abnormalities Prediction \cite{bhandari2023exploring} & LIME,SHAP & Kidney & CT & Accessibility \\

Renal Cell Carcinoma Prediction \cite{han2021explainable} & SHAP & Kidney & CT & Generalisability \\

Electrocardiogram Classification \cite{ganeshkumar2021explainable} & Grad-CAM & Heart & ECG & Data Scarcity \\

Heart Disease Prediction \cite{sreeja2023deep} & Grad-CAM, LIME & Heart & ECG & Accessibility\\

3-lead Electrocardiogram Classification \cite{le2023lightx3ecg} & Grad-CAM & Heart & ECG & Training Complexity\\

Pediatric Heart Transplant Rejection \cite{giuste2023explainable} & Grad-CAM* & Heart & Whole Slide Images & Data Limitation\\

Cardiomegaly Diagnosis \cite{yoo2021diagnosis} & LIME & Heart & X-ray & Data Quality \\

Cardiac Disease Diagnosis \cite{anand2022explainable} & SHAP & Heart & ECG & Practicality \\

Gastric Cancer Diagnosis \cite{chempak2022evaluation} & SM & Stomach & Endoscopic Images & Data Scarcity \\

Gastrointestinal Tract Disorders Classification \cite{nouman2023localization} & Grad-CAM & Stomach & Endoscopic Images & Image Quality\\

Gastric Cancer Classification \cite{mukhtorov2023endoscopic} & Grad-CAM & Stomach & Endoscopic Images & Data Scarcity\\

Gastrointestinal Cancer Classification \cite{auzine2023classification} & SHAP & Gastrointestinal Tract & Endoscopic Images & Practicality \\

Cardiovascular Disease Prediction \cite{guleria2022xai} & SHAP & Cardiovascular & CT, X-ray & Data Scarcity \\

Bladder Cancer Prediction \cite{kirbouga2023bladder} & LIME, Anchor, SHAP & Bladder & Gene Microarrays & Data Scarcity\\

Prostate Cancer Tissue Detection \cite{ramirez2023explainable} & SHAP & Prostate & Gene Expression Data & Practicality\\

Vesicoureteral Reflux Grading \cite{khondker2022machine} & SHAP & Vesicoureteral & Voiding Cystourethrograms & Utility\\

Anterior Disc Displacement Diagnoses \cite{yoon2023explainable} & LRP, IG & Temporomandibular & MRI & Accuracy, Practicality\\

Caries Lesions Detection \cite{ma2022towards} & LRP & Dentistry & Near-infrared Light Transillumination & Practicality\\

Calculus Detection \cite{buttner2023impact} & SHAP & Dentistry & Bitewing Radiographs & Generalisability\\

Alveolar Bone Defect Classification \cite{miranda2023interpretable} & Grad-CAM* & Dentistry & Cone-Beam CT & Human Assistance\\

\bottomrule
\end{tabular}
}
\end{center}
\caption{XAI visual applications in the medical field. Table is sorted by anatomical location. XAI techniques are abbreviated as outlined in Section \ref{tech}. Note that certain techniques followed by a star (*) refer to variants of the original technique(e.\,g., Grad-CAM*). CT: Computed Tomography; MRI: Magnetic Resonance Imaging; OCT: Optical Coherence Tomography; ECG: Electrocardiogram.}
\label{tab:table2}
\end{table*}

Table \ref{tab:table2} presents the most recent XAI visual medical publications published in recent years. Each row displays the name of the medical techniques targeted by the publication, the selected XAI technology, the target anatomical location and the visual modality scenario. In addition, based on our subjective assessment combined with the limitations sections of each paper, where applicable, we extracted the key shortcomings most significantly impacting XAI for each study. Considering space limitations, we selected the representative XAI medical applications based on medical anatomical locations, diversity of XAI techniques,  and publication timeframe to ensure a comprehensive and balanced overview (the selection logic also applies to Table \ref{tab:table3} and Table \ref{tab:table4}).

LIME and SHAP are utilised in \cite{aldughayfiq2023explainable} to generate local and global interpretations of the DL model for improving the diagnosis of retinoblastoma, respectively. The project involves collecting and labelling 400 retinoblastoma and 400 non-retinoblastoma fundus images, dividing them into training, validation, and test sets, and training the models using transfer learning with the pre-trained InceptionV3 model. LIME and SHAP are applied to interpret the model's predictions on the validation and test sets. Both XAI techniques identify typical features of retinoblastoma fundus, such as yellow-white mass, calcification and retinal detachment, which are consistent with clinical observations. And authors announced that this is the first study to apply LIME and SHAP for the task of retinoblastoma detection.

A system is presented in \cite{dipto2023xai} for recognising white blood cell types using a variant of Vision Transformers (ViTs). The training phase shows that the ViTs-based model had a significantly faster learning curve, and the models achieved an accuracy of 83\%-85\%. Additionally, the authors attempt to visualise the attention maps of the Vision Transformer using Grad-CAM. Although the attention maps are unevenly distributed across the image and fail to capture the regions used for final classification, the authors suggest that this could be improved by removing the lowest attentional values through a minimal fusion approach. This indicates potential for improving this method.

A lightweight CNN model is designed to accurately detect various renal abnormalities, including cysts, stones, and tumours in \cite{bhandari2023exploring}. SHAP technique is applied to the CNN model to interpret its predictions across four categories (cyst, normal, stone, tumor). The results show how red and blue pixels highlight important features, with red pixels increasing the probability of a specific prediction and blue pixels decreasing it, thereby providing visual explanations for the model's behaviour. LIME is used to extract numerical features and perturbations, identifying key image regions contributing to specific class predictions. The study also present CT images to medical professionals and then use the results of the model and XAI framework to validate the feedback of medical professionals. The result indicates that the proposed study can be effectively used to identify renal abnormalities in patients, and the explanation generated by the XAI technology were consistent with the medical experts' assessment.

In \cite{deperlioglu2022explainable}, the authors propose an explainable diagnostic method for glaucoma that combines image processing and CAM. The explainability of the method is assessed by 15 physicians. The performance of the explainable CNN model is tested on three retinal image datasets. They first conduct a usability test on the CAM-based CNN model. The conclusion showed that physicians found this system is easy to use in general tasks of glaucoma diagnosis. Subsequently, they compare CAM with alternatives (Grad-CAM and AM), and it is found that CAM has the best overall performance because it has a simple structure when it comes to integration. In conclusion, the authors believe that the introduction of XAI improves the trustworthiness and interactivity in glaucoma diagnosis, but by comparing the results of this study with past studies only based on DL, it is found that the use of XAI may lead to decreased diagnostic performance and reduced accuracy.

The RISE method is optimised in \cite{highton12evaluation} to extend the explanation scope from 2D natural images to 3D medical images. The performance of the method in black-box cerebral haemorrhage classification is verified through an evaluation consistent with clinical interpretability guidelines. The assessment guidelines includes Quality Assessment and Quantitative Assessment. The Quality Assessment follows the framework proposed by \cite{jin2023guidelines} and considers five dimensions: G1: Understandability, G2: Clinical relevance, G3: Truthfulness, G4: Informative plausibility and G5: Computational efficiency to evaluate how faithful the resulting saliency map is to the underlying pathology of interest. The Quantitative Assessment adjusts the indicators in G5 to calculate Hausdorff Distance, Root Mean Square Surface Distance and number of Erroneous Contiguous Regions to achieve the purpose. The RISE method in the paper is applied to a 3D-CT dataset-based brain haemorrhage detection classifier. It is found that using various unit geometries to generate masks improved the saliency heatmap and increased the robustness of complex haemorrhage distributions. The evaluation results, consistent with those of clinicians, underscore the importance of employing various mask geometries to enhance the robustness of medical image datasets, particularly in cases where salient features are unknown. And the authenticity and informativeness of the XAI RISE method are successfully confirmed.

A comprehensive model for lung segmentation is introduced in \cite{teixeira2021impact} to emphasise the importance of diagnosing and explaining COVID-19. A U-Net CNN is employed for semantic segmentation and VGG, ResNet, and Inception for classification. The results indicate that segmentation enhances multi-class and COVID-19 recognition. After the introduction of XAI technology LIME and Grad-CAM, the prediction quality is significantly improved despite a reduction in the $F_1$ score.

There is a well-trained neural network model for classifying three skin tumour types: melanocytic naevi, malignant melanoma, and malignant keratosis \cite{lucieri2020interpretability}. Additionally, a new training method and a new method for saliency testing are presented. The neural network model is trained using a concept activation vector (CAV) that maps the model to a human-understood concept. By utilising TCAV to test the significance of concepts, the authors investigate how classifiers learn and encode human-understandable concepts in their underlying representations. The TCAV scores indicate that the neural network did indeed utilise disease-related concepts expectedly when making predictions. This finding establishes a crucial basis for future development of CAV-based neural network interpretation methods.

In \cite{mertes2022ganterfactual}, an innovative method based on Generative Adversarial Networks (GANs) is introduced, accomplished by creating counterfactual images to provide more acceptable justifications for users.  They conduct a user study to compare CE with the popular LIME and LRP explainability methods. In detail, they choose the explanation of a classifier that distinguishes between X-ray images of lungs that are infected by pneumonia and lungs that are not infected. They then survey respondents based on five dimensions. The evaluation results indicate that CE outperformed the other two methods in the dimensions of mental model, explanation satisfaction, trust, emotion, and self-efficacy. However, there are limitations to this method. For instance, it might dissatisfy some users due to its exaggerated modification of the original features. Additionally, it could not cover scenes that rely more on non-textual information because of its dependence on the texture structure of the image.

The authors in \cite{anand2022explainable} introduce the ST-CNN-GAP-5 model for ECG classification using the PTB-XL dataset, achieving state-of-the-art performance at the time. They integrate SHAP to identify critical ECG waveform segments used for diagnosing heart disorders. These visual explanations are validated by clinicians to ensure the model's explainability and reliability for diagnostic purposes. This approach demonstrated a balanced trade-off between high performance and explainability.

\subsection{Audio XAI Applications in Medicine}

The utilisation of audio data in AI systems has gained considerable traction owing to its inherent expressive capabilities. Particularly in the medical domain, the discernment of pathological changes in the lungs and vocal systems presents promising avenues for diagnostic applications. Noteworthy advancements include the proposal of audio-based AI models tailored for critical medical conditions such as cardiovascular disease \cite{abbas2024artificial, deng2020heart}, COVID-19 \cite{10.3389/fdgth.2022.789980, Coppock356}, Alzheimer's disease \cite{10.3389/fpsyg.2020.624137, liu2020new}, or depression \cite{ringeval2019avec, SARDARI2022116076, min2023detecting}. However, the explainability of these audio AI models remains largely unexplored compared to visual models, raising concerns about their trustworthiness. Thus, enhancing the interpretability of audio AI models targeted for medical purposes emerges as a significant research field to ensure their reliable integration into critical healthcare applications.

Several studies delve into interpreting audio models' predictions through various explainability methods. The classification of lung sounds using a lightweight CNN model is studied in \cite{wanasinghe2024lung} by integrating various audio features such as spectrograms. To enhance the explainability and trustworthiness of the classification model, they incorporate Grad-CAM and SM. Similarly, \cite{lo2022explainable} explores the behaviour of a CNN when classifying the types of respiratory sounds. They also provide a visual explanation of the model by applying Grad-CAM to understand the important features and build reliable diagnostic models for medicine and telemedicine applications.

In \cite{ren2022deep}, a DL model designed to classify heart sounds emphasising explainability is presented. The authors use AM to explore the visualisation of the learnt high level representations of the proposed model which contributes to model interpretability. Focusing on explaining the detection of COVID-19 from audio cough data, CoughLIME \cite{wullenweber2022coughlime} extends the LIME method by integrating Non-negative Matrix Factorisation (NMF) to enhance the interpretability of audio processing models. It decomposes the input audio into components and assigns importance to each component to generate listenable explanations that can be interpreted by humans.

\begin{table*}
\begin{center}
\resizebox{\textwidth}{!}{
\begin{tabular}{cccccccc}%
\toprule
Name & XAI Techniques & Anatomical Location & Modality & Cons \\
\bottomrule
Heart Sound Classification \cite{ren2022deep} & AM & Cardiovascular & Heart Sound Signals & Data Size Limitation \\
Respiratory Disease Detection \cite{wanasinghe2024lung} & Grad-CAM, SM &  Lung & Lung Sound Signals & Performance-explanation trade-off \\
Classification of Respiratory Sound \cite{lo2022explainable} & Grad-CAM & Lung & Respiratory Sounds Signals &  Data Size Limitation \\
COVID-19 Detection \cite{wullenweber2022coughlime} & LIME & Lung & Cough Audio & Generalisability  \\
Discrimination of Vocal Cord Pathology \cite{seedat2020automated} & SHAP & Vocal Cord & Phonation Audio Recordings & Generalisability \\
\midrule%

\bottomrule
\end{tabular}
}
\end{center}
\caption{XAI audio applications in the medical field. XAI techniques are abbreviated as outlined in Section \ref{tech}.}
\label{tab:table3}
\end{table*}

\subsection{Multimodal XAI Applications in Medicine}

Multimodal AI is an emerging paradigm in AI that integrates various data types and multiple outputs, which can imitate human cognition by smoothly processing and integrating various modalities. This integrated method offers a more profound comprehension of the user's requirements, resulting in a more intuitive and richer user experience \cite{stappen2023integrating}. While recent literature \cite{rodis2024multimodal}  has begun to analyse various methods in multimodal XAI, there remains no established, widely accepted taxonomy in the field. In this paper, we attempt to give two categories of multimodal data based on the execution stage: \textit{XAI Multimodal Input} and \textit{XAI Multimodal Output}. 

\textit{XAI Multimodal Input} refers to the model receiving data from multiple sources, (e.\,g., Histological records, medical images, and audio) and then explaining how the model extracts information and makes decisions from the multiple sources by integrating these different modalities.

\textit{XAI multimodal output} is designed to provide a diverse range of explanation outputs, (e.\,g., text, audio and images) that work together to explain the decisions made by the model.

Table \ref{tab:table4} presents the most recent XAI multimodal medical publications published in recent years. Each row displays the name of the medical technology targeted by the publication, the selected XAI techniques, the target anatomical location, the multimodal stage and the shortcomings it faced. 

ExAID \cite{lucieri2022exaid} is a multimodal output explanatory framework designed to provide transparent decision-making for skin lesion diagnosis. This framework utilises TCAV technology and its derivative, Concept Localisation Maps (CLMs), to offer multimodal explanations. TCAV technology has been previously introduced, and CLMs expand upon CAV by localising regions in the latent space of a trained image classifier relevant to learned concepts. By inputting labelled dermatoscopic images into a DL model trained with these XAI techniques, ExAID delivers accurate diagnostic outcomes accompanied by detailed textual explanations and visual representations. This framework supports clinical diagnostic interfaces and educational models for research and training purposes.

In \cite{guarrasi2023multi}, the authors propose a novel joint fusion technique to optimise the combination of multimodal deep architectures by maximising the classification evaluation metric and the diversity measure among learners. The authors employ this methodology to utilise both X-ray and patient clinical data as inputs, applying Pareto multi-objective optimisation to consider the performance indicators and diversity scores of multiple candidate single-modal neural networks to predict the severity of patients infected with COVID-19. The findings of the research demonstrate that this method yields considerable improvements in prediction performance. Furthermore, the authors integrate IG and Grad-CAM technology into the joint fusion technology, elucidate the rationale behind the model's decision-making process, and illustrate the role of each modality in the decision-making process and its relative contribution.

\cite{dentamaro2024enhancing} introduce a multimodal DL framework for early Parkinson's detection, integrating MRI with clinical data. They apply DenseNet, ResNet, and Vision Transformer (ViT) models to MRI, while processing clinical data through an Excitation Network (EN). A joint co-learning strategy fuses these modalities, enhancing predictive accuracy. IG and AM are employed to improve explainability. DenseNet highlights regions associated with cognitive and motor decline, while ViT focuses on lateral ventricles, aiding biomarker identification, and decision-making clarity.

\begin{table*}
\begin{center}
\resizebox{\textwidth}{!}{
\begin{tabular}{cccccccc}%
\toprule
Name & XAI Techniques & Anatomical Location & Multimodal Stage & Multimodal 
 Modality & {Cons}\\
\midrule
Skin Lesions Diagnosis \cite{lucieri2022exaid}& TCAV*& Skin & Output& Dermoscopic images, Texts  & Data Quality  \\

Parkinson's detection \cite{dentamaro2024enhancing}& IG, AM& Brain & Input & MRI, clinical data & Data Scarcity \\

Alzheimer's Prediction  \cite{rahim2023prediction}& Grad-CAM*  & Brain & Input &  MRI with Multi-Biomarkers  & Data Compatibility \\ 

Alzheimer's Diagnosis \cite{odusami2023explainable}   & SHAP  & Brain & Input & MRI, PET &Practicality \\ 

Parkinson’s Disease Detection \cite{junaid2023explainable}& LIME,SHAP  & Brain & Input &  Patient characteristics, Biosamples, Medication History & Generalisability\\ 

Hand Gesture Recognition \cite{kang2022reduce}   & SM,IG,SHAP  & Muscle & Input &IMU, EMG & Explanation consistency \\ 

Pleural effusion analysis \cite{da2022biomedical}  & LRP, Grad-CAM, LIME & Chest & Output & X-Ray Saliency map, Texts  & Data Dependency\\ 

Chest X-Ray Classification \cite{ketabi2023multimodal} & Grad-CAM  & Chest & Input &  X-ray, Radiology reports, Eye-Gaze  &  Redundancy\\ 

COVID-19 Prediction \cite{guarrasi2023multi} & IG,Grad-CAM  & Chest & Input &  X-ray, Clinical Information  & Generalisability \\ 

COVID-19 Diagnosis \cite{saif2021capscovnet}  & Grad-CAM  & Chest & Input &  X-ray, Ultrasound  & Data Scarcity \\ 

COVID-19 Diagnosis \cite{rahman2021multimodal}  & SHAP, LIME, Grad-CAM  & Respiratory System & Input & Radiological Media   & Data Sacarcity\\ 

\bottomrule
\end{tabular}
}
\end{center}
\caption{XAI multimodal application in the medical field. Table sorted by anatomical location. XAI techniques are abbreviated as outlined in Section \ref{tech}. Note that certain techniques followed by a star (*) refer to variants of the original technique  (e.\,g., Grad-CAM*). MRI: Magnetic Resonance Imaging; PET: Positron Emission Tomography; IMU: Inertial Measurement Unit; EMG: Electromyography.}
\label{tab:table4}
\end{table*}

\section{Discussion}
\label{discuss}
In this section, we discuss the observed patterns and challenges identified after evaluating the XAI criteria taxonomy, techniques, and their applications in the medical field.
\subsection{Observation on Criteria Taxonomy and Techniques }

The evaluation shows that the majority of XAI techniques used in the medical field are perception-based. These methods provide intuitive, visually or aurally accessible insights that clinicians can easily interpret. This aligns with clinical needs, as visual methods facilitate a quick understanding by mapping model outputs directly to anatomical features. In contrast, mathematically grounded approaches, though theoretically rigorous, may not offer the same immediacy or intuitive appeal in practical healthcare settings, potentially limiting their clinical applicability.

The majority of XAI techniques currently used in medical AI applications are model-specific, particularly those based on gradient and backpropagation methods. This trend aligns with the prevalence of DL in the medical field, which relies on multi-layered structures and gradient-based optimisation. Gradient and backpropagation-based XAI techniques leverage these characteristics to provide feature attributions and detailed insights. As traditional ML models lack the complex, differentiable structures of DL, these XAI techniques are less suited to them.

The evaluation shows that most XAI techniques belong to the post-hoc category. They can explain a model after it has completed its inference, without requiring any modification of the model's structure or retraining. This provides considerable flexibility and broad applicability for specific trained models. This pattern is similar to medical judgment, where doctors typically make inferences and interpretations based on the information obtained after completing examinations or receiving test results. However, this approach limits a comprehensive understanding of the models' dynamic decision-making process.

In addition, among the four taxonomy criteria, although some techniques can provide both local and global explanations, they are mutually exclusive in other criteria. This suggests that no single XAI technology is comprehensive. Therefore, some researchers propose using multiple XAI technologies or variants in combination within the same model to integrate their respective advantages and achieve a more comprehensive explanation effect \cite{toussaint2024explainable, 12a6e929131544c18c146d2a90337631}. However, most of the literature still tends to rely on a single XAI technique for evaluation, which may result in an incomplete understanding of model explainability.

Besides, in the review of XAI research, a persistent challenge is the delineation of specific methodologies within the scope of XAI techniques. This issue stems from the absence of universally accepted definitions or standards for categorising particular techniques (such as AM) within the XAI domain, often leading to subjective assessments by researchers. Furthermore, numerous studies attempt to provide explanations for AI models without formally integrating established XAI frameworks or techniques, which may lead to omissions in statistical analysis while reviewing.

\subsection{Observation from Medical Applications}
Based on an evaluation of XAI applications in the healthcare field, several common challenges emerge.

In terms of data, many XAI studies rely heavily on publicly available datasets, leading to significant variations in data quality. The lack of direct collaboration with medical institutions limits feasibility assessments in real clinical settings. This dependency on public data also raises concerns about privacy and security, undermining the practicality and applicability of research findings. Additionally, data scarcity is a persistent issue, particularly in cases where disease-specific data is limited, further hindering the model’s ability to generalise to complex cases and diverse clinical scenarios.

Time and resource demands present significant barriers to XAI’s practical implementation. Many XAI techniques require substantial computational resources and time for data processing and explanation generation. While these methods enhance model explainability, they also escalate costs and increase processing time. Many studies have highlighted the trade-off between explainability and performance, which is especially critical in clinical environments where real-time decisions are essential. This resource-intensive nature of XAI integration often discourages researchers from embedding these techniques into primary tasks.

Generalisability remains a shared challenge for both XAI techniques and primary tasks. Existing XAI methods frequently struggle with stable transferability across diverse settings or model structures. In healthcare, varying equipment, data formats, and individual patient characteristics between institutions restrict XAI’s applicability and consistency, ultimately impacting its broader use in medical environments.

In terms of user needs and applicability, XAI applications face numerous challenges in real-world settings. Many XAI applications have not achieved end-to-end explanations, remaining in experimental stages and thus lacking evaluation in actual medical contexts. Additionally, XAI methods often lack consistency in explanations, where different techniques can produce contradictory conclusions, necessitating human expert intervention. This reliance on experts not only complicates the system but also undermines user trust. Such inconsistency in explanations increases the workload on healthcare professionals, potentially affecting diagnosis and treatment accuracy, and further restricting XAI’s broader adoption.

Assessments indicate that there are currently few XAI applications that thoroughly explore the use of medical technology from the perspectives of fairness and bias. This gap has resulted in significant shortcomings in key areas such as data bias, algorithm transparency, and patient privacy protection. Most existing XAI applications primarily focus on improving model explainability and performance while neglecting the potential unfair impacts on vulnerable groups and the associated ethical risks. Furthermore, the lack of targeted, diverse datasets contributes remarkably to these issues.

The number of available publications suggests that XAI research in the audio and multimodal healthcare fields is comparatively less abundant than that in CV. CV as a foundational AI branch, has established a solid research base and extensive data resources in the medical field, attracting remarkable investment \cite{elyan2022computer}. By contrast, audio and multimodal XAI are relatively new, facing challenges from a lack of historical data and established frameworks. The complexity of audio, particularly its temporal nature, and the intricate relationships between data modalities make preprocessing and feature extraction more challenging. Multimodal approaches usually require integrating two or more modalities, applying multiple XAI techniques, and involving extensive experimentation and consolidation. This process can be time-consuming and often yields suboptimal results. This discourage many researchers from pursuing this line of study.

However, even in the relatively mature CV field, XAI still faces unique challenges in medical applications. First, different imaging techniques target specific anatomical details \cite{hussain2022modern}. For instance, CT highlights bone structures while MRI focuses on soft tissue. XAI must adapt to these differences to provide clinically meaningful interpretations. This customisation requirement implies that a single XAI method cannot cater to all imaging types, necessitating additional fine-tuning and optimisation during technology transfer. Furthermore, the high-dimensional nature of medical imaging greatly increases explainability complexity, especially for 3D images, where the dimensionality challenges grow exponentially compared to 2D images. Not only do 3D images include more pixels and voxels, but they also require XAI to account for intricate spatial relationships and layered structures, which can impede comprehensive and detailed explanations. While current XAI methods perform well with 2D images, applying them to 3D data often demands further adjustments and may not yield equally satisfactory results.

The explainability evaluation of XAI methods often relies on overviews or subjective judgments, which may fail to effectively capture the specific strengths and limitations of XAI tools in various medical tasks. In most works, while performance metrics, such as accuracy or F1 scores, offer precise, quantifiable measures, explainability evaluations often rely on general terms such as “enhancing transparency,” “ensuring fairness,” or “improving traceability,” without clear, standardised quantitative benchmarks specific to medical applications. This non-quantitative approach challenges systematic comparison, further relegating XAI methods to a supplementary position in medical research tasks.   Although frameworks such as the Co-12 properties proposed in \cite{nauta2023anecdotal} provide a foundation for quantitative assessment, the adoption of such systematic evaluation methods remains limited, especially within medical contexts. The QXAI framework proposed by \cite{shaik2023qxai} offers a quantitative explainability approach for patient monitoring systems. However, its high computational complexity limits widespread adoption.

In general, the development and application of XAI technology is a complex process involving multiple levels of technology, ethics, and practical application. To fully realise the potential of XAI in the medical field,  it is necessary to have multi-disciplinary collaboration.  This collaboration is required to develop interpretive AI systems that are both efficient and ethical.

\section{Outlook}
\label{out}
Considering the existing challenges and limitations in the field of medical applications of XAI, some suggestions for future research and development are given:

Given the limitations of current XAI technology, it is recommended that future XAI research should prioritise the development of customised XAI techniques tailored to specific medical needs, offering detailed explanations of medical processes and decision-making logic. For example, XAI methods could be adapted to specific data modalities, such as imaging data in radiology or audio data in the analysis of respiratory diseases, as well as to particular clinical metrics, like tumour progression in oncology or biomarker analysis in personalised medicine. Furthermore, medical researchers should consider integrating multiple XAI techniques to provide a more comprehensive and nuanced understanding of healthcare data. 

To address the challenges of defining and standardising XAI methodology, it is recommended that authoritative institutions or academic organisations collaborate to establish a unified set of definitions and terminology, which helps to clearly define the scope of XAI and standardise its application. After addressing the standardisation concerns, Multi-Criteria Decision Analysis (MCDA) \cite{muhlbacher2016making} can be introduced as a quantitative evaluation method, that can effectively resolve the uncertainty of evaluation standards. As an advanced decision support technology, MCDA enables comprehensive evaluation and comparison of different solutions or models under multiple criteria. The MCDA method can be used to integrate explainability-related evaluation dimensions with accuracy dimensions to achieve a unified evaluation method. This comprehensive evaluation framework is particularly suitable for fields such as medical care that require a high degree of accuracy and explainability.

To address the high demands for explainability and safety in healthcare, more rigorous explainable AI methods, such as Formal XAI, can be introduced \cite{marques2022delivering, bassan2023towards}. Unlike traditional XAI approaches, Formal XAI relies on mathematical and logical reasoning through formal verification to produce verifiable explanations, rather than on heuristic or empirical examples. This approach offers greater rigour in explanations. Future research should focus on enhancing its computational efficiency and exploring its potential applications in high-risk areas like healthcare, ultimately improving the reliability and explainability of AI models in clinical settings.

Given the current reliance on publicly available datasets in research, future strategies should focus on establishing tighter partnerships with healthcare organisations. By establishing a secure data-sharing platform based on cloud computing technology, such as the Federated Data Platform (FDP) being developed by NHS England \cite{dunn2022using}, region healthcare systems can provide a robust guarantee of data privacy and security while facilitating real-time data sharing across different organisations. This allows research to rely on broader and more realistic medical data, significantly improving the accuracy and depth of research. In addition, when conducting XAI research, it is important to encourage active participation from both patients and doctors. This will improve the quality and relevance of data collection, as well as enhance the practicality and social acceptance of the research results and explanations given.

As mentioned ahead, while some XAI methods have shown great promise in explainability, their application in healthcare remains limited \cite{petsiuk2018rise, achtibat2022towards, o2020generative}. These methods are often validated in other fields but lack thorough exploration in medical contexts. Future research could delve into their potential in healthcare, possibly introducing new explainability benefits for medical AI systems. Additionally, integrating these emerging methods with existing mainstream XAI techniques in healthcare could create a more comprehensive framework, enhancing models' clinical feasibility and applicability.

As the field of P4 medicine (personalised, predictive, preventive, and participatory) gains prominence \cite{flores2013p4}, it is imperative that XAI systems evolve to support model personalisation and enhanced interactivity within the healthcare context. As medical models become increasingly patient-specific, XAI should facilitate explanations that adapt to the nuances of individualised treatment plans. Furthermore, the interactive nature of these systems is important for enabling clinicians to actively engage with and refine model outputs in real time. By offering adaptive, context-aware explanations, XAI will be better equipped to meet the demands of complex, personalised care and support more precise decision-making in healthcare context.

To address ethical and bias issues, future developments in XAI should focus on specific technical and application improvements. Firstly, It is necessary to design more representative datasets that include diverse populations across different races, genders, and ages, particularly where disease manifestation and treatment outcomes vary among individuals \cite{baird2020considerations}. In cases where data is scarce, data augmentation or generative techniques can be employed to supplement existing datasets \cite{juwara2024evaluation}. Additionally, introducing fairness assessments during the data collection phase can effectively reduce sources of bias. Secondly, fairness metrics (such as disparate impact and equal opportunity) should be applied during model training and interpretation to detect and mitigate potential biases \cite{chen2023algorithmic, sun2024audiobasedkinshipverificationusing}. These measures will enhance the fairness and ethical standards of XAI in the healthcare sector.

Considering how far audio medicine and multimodal medicine have lagged in XAI research, future research directions should focus on deeper excavation and development of audio medicine and multimodal medicine in XAI research. This involves utilising innovative methods to manage the time series characteristics of audio data and the intricate interplay between multimodal data. In addition, research should explore innovative methods of data explanation, such as sonification \cite{schuller2021towards}. Sonification is a technique that transforms complex data into non-verbal sound forms and combines them with other modalities such as text and images. This provides a new way of explaining and understanding healthcare data.

\section{Conclusions}
\label{con}
In this review, the background of XAI was explored, and a comprehensive collection and integration of applications in the medical field in recent years was conducted. The importance of explainability in medical scenarios was emphasised from the perspectives of patients and doctors, followed by the definition of XAI and related terms. We consolidated four existing taxonomy criteria to provide a comprehensive framework of 19 different XAI techniques. We also utilised a implementation-based approaches to classify these XAI technologies and evaluate each technology based on the aforementioned taxonomy criteria. Furthermore, a multitude of XAI application cases in the medical field spanning the visual, audio, and multimodal domains were gathered. By evaluating and analysing these techniques, we clarified the current limitations of XAI in the medical field and proposed corresponding perspectives. Finally, we concluded with some profound insights that provide valuable guidance for future development. In particular, future XAI needs to be more multimodal, personalised, and extensively tested with the target users to lead into a genuinely trustworthy future usage of AI in medicine.



\bibliographystyle{ACM-Reference-Format}
\bibliography{sample-base}

\end{document}